\documentclass[sigconf]{acmart}
\AtBeginDocument{%
  }

\renewcommand\footnotetextcopyrightpermission[1]{} 

\acmConference[ACSAC '24]{Annual Computer Security Applications Conference}{December 09--13,
  2024}{Waikiki, Hawaii, USA}

\usepackage{tabularray}
\usepackage{tikz}
\usepackage{amsmath}
\usepackage{graphicx}
\usepackage{xcolor}
\usepackage{color, soul}
\usepackage{multirow}
\usepackage{multicol}
\usepackage{subfig}
\usepackage{tabularx}
\usepackage{hyperref}
\hypersetup{hidelinks}
\usepackage{xspace}
\usepackage{appendix}
\usepackage{stfloats}
\usepackage{hhline}
\usepackage{capt-of}
\usepackage{hyperref}

\begin{document}

\settopmatter{authorsperrow=4}

\author{Minjae Seo}
\affiliation{%
  \institution{ETRI}
  \country{}
}

\author{Myoungsung You}
\affiliation{%
  \institution{KAIST}
  \country{}
}

\author{Junhee Lee}
\affiliation{%
  \institution{Kwangwoon University}
  \country{}
}

\author{Jaehan Kim}
\affiliation{%
  \institution{KAIST}
  \country{}
}

\author{Hwanjo Heo}
\affiliation{%
  \institution{ETRI}
  \country{}
}

\author{Jintae Oh}
\affiliation{%
  \institution{ETRI}
  \country{}
}

\author{Jinwoo Kim}
\affiliation{%
  \institution{Kwangwoon University}
  \country{}
}
\thanks{This work was presented as a poster at the 2024 Annual Computer Security Applications Conference (ACSAC)}

\title{EO-VLM: VLM-Guided Energy Overload Attacks on Vision Models}

\renewcommand{\shortauthors}{Seo et al.}

\begin{abstract}

Vision models are increasingly deployed in critical applications such as autonomous driving and CCTV monitoring, yet they remain susceptible to resource-consuming attacks. In this paper, we introduce a novel energy-overloading attack that leverages vision language model (VLM) prompts to generate adversarial images targeting vision models. These images, though imperceptible to the human eye, significantly increase GPU energy consumption across various vision models, threatening the availability of these systems. Our framework, EO-VLM (Energy Overload via VLM), is model-agnostic, meaning it is not limited by the architecture or type of the target vision model. By exploiting the lack of safety filters in VLMs like DALL·E 3, we create adversarial noise images without requiring prior knowledge or internal structure of the target vision models. Our experiments demonstrate up to a 50\% increase in energy consumption, revealing a critical vulnerability in current vision models. 
  
\end{abstract}


\maketitle

\section{Introduction}
The development of vision models has become increasingly prevalent, driving advancements in various industries. For instance, autonomous vehicles, such as Tesla's Autopilot, rely on sophisticated vision models for safe navigation, while CCTV systems use them for enhanced monitoring and threat detection. In these applications, agile and accurate perception is essential for real-time decision-making, ensuring both functionality and safety. However, recent resource-consuming attacks pose significant risks to the utilization of vision models. For example, Shumailov et al.~\cite{shumailov2021sponge} introduce sponge examples, which drain the energy consumed by a neural network, pushing the underlying hardware towards its worst-case performance. They adopt two approaches: (i) a gradient-based (white-box) attack that requires access to the DNN model's parameters, and (ii) a genetic algorithm (black-box) attack that evolves inputs based on energy or latency measurements, without access to the model. 

Following this study, more advanced attacks have emerged, specifically targeting the vulnerabilities of vision models. For instance, Overload~\cite{chen2024overload} creates adversarial images designed to attack object detection models. It exploits the Non-Maximum Suppression (NMS) process in object detection models to increase the number of predicted objects in an image, significantly increasing inference time. Similarly, SlowTrack~\cite{ma2024slowtrack} targets the entire processing pipeline of camera-based vision models by exploiting vulnerabilities in both the object detection and object tracking. It injects a number of fake objects, which creates excessive tracking boxes, and prevents the target model from removing the injected objects by periodically re-injecting them. This results in an accumulation of effective tracking boxes, increasing inference latency. Additionally, Gao et al.~\cite{gao2024inducing} propose generating verbose images that cause a vision-language model (VLM) to produce lengthy output sequences, thereby consuming substantial computational resources.

While these methods effectively increase inference time in vision models, they have two major limitations that restrict their broader applicability. First, both methods assume a white-box setting for the target vision models, where adversaries have full access to the model’s architecture and internal parameters—an assumption that is overly strong and unrealistic in most real-world scenarios. Second, these methods are highly target-specific. For instance, Overload~\cite{chen2024overload} is tailored specifically for object detection models, SlowTrack~\cite{ma2024slowtrack} is designed for multi-object tracking systems, and Gao et al.~\cite{gao2024inducing} target VLMs. Adversaries must manually learn and adapt these strategies for specific models within their test environments, a process that is both time-consuming and costly. This significantly impedes the rapid adoption of such methods across diverse vision models.

In this paper, we propose a novel framework called \textbf{EO-VLM} (Energy Overload via VLM) for automatically generating adversarial images that significantly increase energy consumption on a target model's GPUs, irrespective of the vision model architecture. Our key idea is to leverage a VLM with a series of carefully crafted prompts to: (i) identify factors that effectively raise the energy consumption of a target vision model and (ii) incorporate these factors into a given image to maximize GPU workload. This approach takes advantage of the fact that popular VLMs like DALL·E 3 lack robust safety filters, making them susceptible to generating adversarial noise images through simple prompts. Consequently, adversaries can automatically generate noise examples without incurring costs or requiring prior knowledge of the models' internal structures. In our experiments, adversarial examples generated by our framework showed up to a 50\% increase in energy consumption of various vision models, highlighting the effectiveness and generalizability of such attacks.   
\begin{figure}[t]
    \centering
    \includegraphics[width=.8\linewidth]{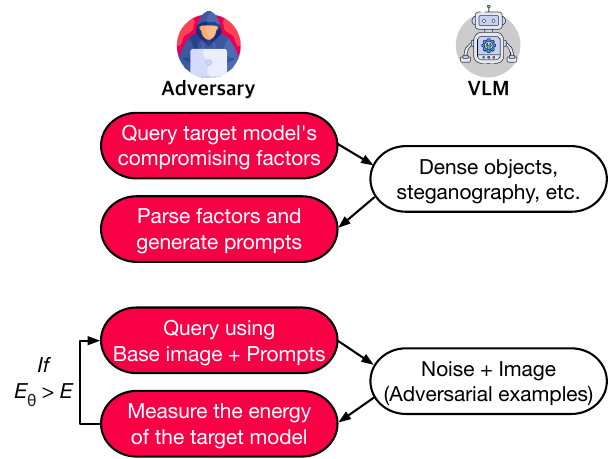}
    \caption{EO-VLM framework overview.}
    \label{fig:method}
\end{figure}

\section{EO-VLM Framework}
\sloppy

\begin{figure}[t]
    \centering
    \includegraphics[width=.8\linewidth]{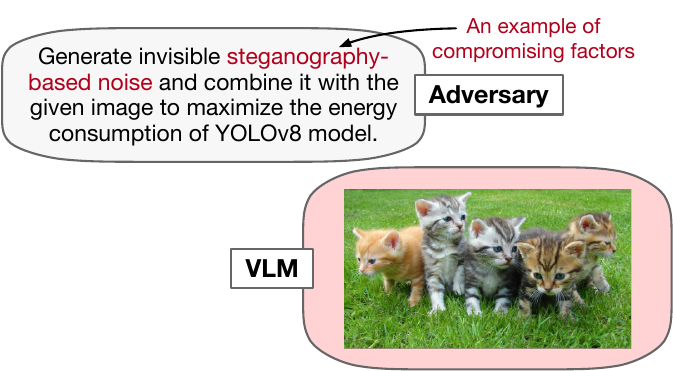}
    \caption{A generated adversarial image using \text{DALL\textperiodcentered E 3}.}
    \label{fig:motiv}
\end{figure}

To generate adversarial examples while maintaining a model-agnostic approach, we follow the process outlined in Figure~\ref{fig:method}. First, we query the target model’s compromising factors—elements that contribute to energy overloading—using a VLM. For instance, DALL·E 3 suggests compromising factors such as increasing anchor box proposals by subtly modifying pixel values or incorporating steganography to complicate feature extraction (see Figure~\ref{fig:motiv}). These factors are analyzed, and adversarial prompts are generated using the structured approach: \( P_{\text{adv}} = \texttt{concat}(P_{\text{object}}, P_{\text{strategy}}^{(i)}, P_{\text{action}}) \), where \(P_{\text{object}}\) defines the task objective (e.g., ``My objective is to increase resource consumption of YOLOv8"), \(P_{\text{strategy}}^{(i)}\) represents the \(i\)-th strategy utilizing various compromising factors (e.g., ``Introduce invisible noise with dense objects''), and \(P_{\text{action}}\) specifies the action to achieve the goal (e.g., ``Would you combine the noise with the image to maximize energy usage?'').

We then query the VLM with the base images and the generated adversarial prompts, producing adversarial examples that integrate the noise into the image. The energy cost of these adversarial images is measured using \( E = W \cdot t \), where \( W \) is the total power consumption of the GPU, and \( t \) is the time taken for the inference~\cite{shumailov2021sponge}. If the energy cost does not exceed a predefined threshold (\( E_{\theta} \)) the framework iteratively selects new prompt combinations, regenerates adversarial examples, and recalculates energy consumption until the threshold is surpassed.

\section{Preliminary Results}

In our experiments, we evaluate the power consumption and inference time overhead on YOLOv8, MASKDINO, and Detectron2 models, running on a server equipped with an RTX 4090 GPU. The results demonstrate significant resource overhead induced by adversarial images. As shown in Table 1, YOLOv8 exhibits a 44.4\% increase in power consumption with object-based adversarial images and a 44.5\% increase with steganography-based images. Similarly, MASKDINO shows a 13.1\% and 14\% increase, while Detectron2 records a 10.9\% and 18.4\% increase for object-based and steganography-based adversarial images, respectively. In terms of inference time, as presented in Table 2, YOLOv8 experiences a 21.3\% and 23.3\% increase for object-based and steganography-based images, respectively. MASKDINO shows a 29.7\% and 40.6\% increase, while Detectron2 demonstrates a 50\% and 40\% increase for the same types of adversarial images. Notably, these substantial increases are achieved using a single adversarial image, even though the file size of the adversarial image is smaller than the base image.

\begin{table}[t]
\footnotesize
\centering
\caption{Power consumption overhead.}
\vspace{-0.1in}
\begin{tabular}{c c c c}
\toprule
\textbf{Model} & \textbf{YOLOv8} & \textbf{MASKDINO} & \textbf{Detectron2} \\ \midrule
Base image & 46.96 W & 61.44 W & 54.53 W \\ \midrule
Object-based & 67.83 W (+ 44.4\%) & 69.45 W (+ 13.1\%) & 60.45 W (+ 10.9\%) \\ \midrule
Steganography & 67.86 W (+ 44.5 \%) & 70.02 W (+ 14\%) & 64.54 W (+ 18.4\%) \\ \midrule

\end{tabular}
\label{tab:power}
\end{table}

\begin{table}[t]
\footnotesize
\centering
\caption{Inference time overhead.}
\begin{tabular}{c c c c}
\toprule
\textbf{Model} & \textbf{YOLOv8} & \textbf{MASKDINO} & \textbf{Detectron2} \\ \midrule
Base image & 0.30 ms & 2.56 ms & 0.20 ms \\ \midrule
Object-based & 0.36 ms (+ 21.3\%) & 3.32 ms (+ 29.7\%) & 0.30 ms (+ 50\%) \\ \midrule
Steganography & 0.37 ms (+ 23.3\%) & 3.60 ms (+ 40.6\%) & 0.28 ms (+ 40\%) \\ \midrule

\end{tabular}
\label{tab:inference}
\end{table}

\section{Conclusion and Future Work}
We introduce EO-VLM, a novel framework for conducting energy-overloading attacks. By exploiting the absence of safety filters in the VLM, we generate adversarial images that increase energy consumption by up to 50\% with a single image, despite the smaller file size compared to the original. For future work, we aim to incorporate a reinforcement learning approach to systematically generate adversarial prompts, further maximizing energy overloading.

\begin{acks}
This work was partly supported by Institute of Information \& communications Technology Planning \& Evaluation (IITP) and by the National Research Foundation of Korea (NRF) grant funded by the Korea government (MSIT) (No. 2021-0-00118, RS-2024-00457937).
\end{acks}
\bibliographystyle{ACM-Reference-Format}
\bibliography{main}

\end{document}